\documentclass[11pt,a4paper]{article}
\usepackage[hyperref]{emnlp2020}
\usepackage{times}
\usepackage{latexsym}

\aclfinalcopy 

\usepackage{microtype}

\usepackage{graphicx}
\usepackage{amsmath}
\usepackage{amsthm}
\usepackage{booktabs}
\usepackage{algorithm}
\usepackage{algorithmic}
\usepackage{relsize}
\usepackage{adjustbox}
\usepackage{multirow}
\usepackage[font=footnotesize,subrefformat=parens,labelformat=parens]{subfig}
\usepackage{wasysym}

\usepackage{xspace}
\usepackage{enumitem}

\newcommand{\model}[1]{\text{#1}\xspace}
\newcommand{\mdattn}{\model{BiLSTM$+$A}}
\newcommand{\mdbi}{\model{BiLSTM}}
\newcommand{\mdcnn}{\model{CNN}}
\newcommand{\mdbert}{\model{BERT}}

\newcommand{\dataset}[1]{\texttt{#1}\xspace}
\newcommand{\dsyelpsmall}{\dataset{yelp50}}
\newcommand{\dsyelplarge}{\dataset{yelp200}}

\newcommand{\dsrt}{\dataset{rt}}
\newcommand{\dsglove}{\dataset{glove.840B.300d}}

\newcommand{\method}[1]{\textsc{#1}\xspace}
\newcommand{\deepfool}{\method{DeepFool}}
\newcommand{\fgm}{\method{FGM}}
\newcommand{\fgvm}{\method{FGVM}}
\newcommand{\tsai}{\method{TYC}}
\newcommand{\hotflip}{\method{HotFlip}}
\newcommand{\textfooler}{\method{TextFooler}}

\newcommand{\metric}[1]{\text{#1}\xspace}
\newcommand{\accuracy}{\metric{ACC}}
\newcommand{\bleu}{\metric{BLEU}}

\newcommand{\acceptability}{\metric{ACPT}}
\newcommand{\use}{\metric{SEM}}

\newcommand{\secref}[2][]{Section#1~\ref{sec:#2}}

\newcommand{\tabref}[2][]{Table#1~\ref{tab:#2}}
\newcommand{\figref}[2][]{Figure#1~\ref{fig:#2}}

\title{Elephant in the Room: An Evaluation Framework for \\Assessing
Adversarial Examples in NLP}

\author{Ying Xu\\
  IBM Research \\Australia \\
  \\\And
  Xu Zhong \\
  IBM Research \\Australia \\
   \\\And
  Antonio Jose Jimeno Yepes \\
  IBM Research \\Australia \\
   \\\And
  Jey Han Lau \\
  University of \\Melbourne \\
   \\
  }

\date{}

\begin{document}
\maketitle
\begin{abstract}
An adversarial example is an input transformed by small perturbations
that machine learning models consistently misclassify. While there are a
number of methods proposed to generate adversarial examples for text
data, it is not trivial to assess the quality of these adversarial examples, 
as minor perturbations (such as changing a word in a sentence) can lead 
to a significant shift in their meaning, readability and classification 
label.  In this paper, we propose an evaluation framework consisting of 
a set of automatic evaluation metrics and human evaluation guidelines, 
to rigorously assess the quality of adversarial examples based on the 
aforementioned properties.  We experiment with six benchmark attacking 
methods and found that some methods generate adversarial examples with 
poor readability and content preservation.  We also learned that 
multiple factors could influence the attacking performance, such as the 
length of the text inputs and architecture of the classifiers.

\end{abstract}

\section{Introduction}
\label{sec:intro}

\textit{Adversarial examples}, a term introduced in
\newcite{Szegedy+:2013}, are inputs transformed by small perturbations 
that
machine learning models consistently misclassify. The experiments are
conducted in the context of computer vision (CV), and the core idea is
encapsulated by an illustrative example: after imperceptible noises are
added to a panda image, an image classifier predicts, with high
confidence, that it is a gibbon. Interestingly, these adversarial
examples can also be used to improve the classifier --- either as
additional training data \cite{Szegedy+:2013} or as a regularisation
objective \cite{Goodfellow+:2014} --- thus providing motivation for
generating effective adversarial examples.


The germ of this paper comes from our investigation of adversarial
attack methods for natural language processing (NLP) tasks,
e.g.\ sentiment classification, which drives us to quantify what is an 
``effective'' or ``good'' adversarial example. In the context of images, 
a good adversarial example is typically defined according to two criteria:

\begin{enumerate}[label=(\alph*)]
\setlength\itemsep{-0.5em}
\item it has successfully fooled the target classifier;
\item it is visually similar to the original example.
\end{enumerate}

In NLP, defining a good adversarial example is a little more involving. 
While criterion (b) can be measured with a comparable text
similarity metric (e.g.\ BLEU or edit distance) and semantic similarity
metrics (e.g. cosine distance between sentence embeddings), an 
adversarial example should also:

\begin{enumerate}[label=(\alph*),start=3]
\setlength\itemsep{-0.5em}
\item be fluent or natural;
\item preserve its original label.\footnote{In the CV example, if the 
perturbed panda image {looks} like a panda, it fulfils criterion (b) 
and (d). In an NLP task such as sentiment classification, even though a 
perturbed sentence may \textit{look} similar to the original and so 
satisfies criterion (b), the perturbed sentence might have the opposite 
sentiment because of a word change (e.g.\ from \textit{good} to 
\textit{tolerable}).}
\end{enumerate}

These two additional criteria are generally irrelevant for images, as adding minor
perturbations to an image is unlikely to: (1) create an uninterpretable
image (while changing one word in a sentence can render a sentence
incoherent), and (2) change how we perceive the image, say
from seeing a panda to a gibbon (but a sentence's sentiment can be
reversed by simply adding a negative adverb such as \textit{not}).
Without considering criterion (d), generating adversarial examples in 
NLP would be trivial, as the model can learn to simply replace a positive
adjective (\textit{amazing}) with a negative one (\textit{awful}) to 
attack a sentiment classifier, or substitute a numeric token with 
another number to attack a machine comprehension system that is 
queried for the year of an event. In other words, while criterion (d) is 
directly implied by criterion (b) in CV (a visually similar perturbed 
image generally preserves its original label), this is not the case for 
NLP.  To the best of our knowledge, most studies on adversarial example
generation in NLP have largely ignored these additional criteria  
\cite{wang2019robust,ebrahimi2017hotflip,tsai2019adversarial,gong2018adversarial}.


The core contribution of our paper is to introduce a systematic 
evaluation framework that combines automatic metrics and human 
judgements to assess the quality of adversarial
examples for NLP. We focus on sentiment classification as the target
task, as it is a popular application that highlights the importance of
the criteria discussed above. It is worth noting, however, that our 
framework is generic and applies to any NLP task.

We test our evaluation framework on a number of attacking methods, 
ranging from white-box to black-box attacks for generating adversarial 
examples.\footnote{White-box
 attack assumes full access to the target classifier's architecture
and parameters; black-box attack, on the other hand, does not. } For the 
human judgements,  we  crowdsource the annotations to assess criteria 
(b), (c) and (d).  Our results reveal that examples generated from most 
attacking methods are successful in fooling the target classifiers, but 
their language is often unnatural and the original label is not properly 
preserved. We also found that a number of external factors have a 
substantial impact on the attacking performance, such as the length of 
text inputs and the classifier architectures.  Lastly, we evaluate the 
transferability of the adversarial examples and the computational time of 
different attacking methods.  Transferability measures 
how effective the adversarial examples (generated for one classifier) 
are in attacking other classifiers. 

\section{Related Work}
\label{sec:related}

Most adversarial attack methods for text inputs are derived from methods 
originally designed for image inputs. These methods can be categorised 
into three types: gradient-based attacks, optimisation-based attacks and 
model-based attacks.

Gradient-based attacks are white-box attacks
that rely on the gradients of the target classifier with respect to the 
input representation. This class of attacking methods 
\cite{kurakin2016adversarial,dong2018boosting,kurakin2016adversarial} 
are by and large inspired by the fast gradient sign method (FGSM) 
\cite{Goodfellow+:2014}, and it has been shown to be effective in 
attacking CV classifiers. However, these gradient-based methods could 
not be applied to text directly because perturbed word embeddings do not 
necessarily map to valid words. Other methods such as DeepFool 
\cite{moosavi2016deepfool} that rely on perturbing the word embedding 
space face similar roadblocks.  \newcite{gong2018adversarial} propose to 
use nearest neighbour search to find the closest word to the perturbed 
embedding.

Both optimisation-based and model-based attacks treat adversarial attack 
as an optimisation problem where the constraints are to maximise the 
loss of target classifiers and to minimise the difference between 
original and adversarial examples. Between these two, the former uses 
optimisation algorithms directly; while the latter trains a seperate 
model to generate the adversarial examples and therefore involves a 
training process. Some of the most effective attacks for images are 
achieved by optimisation-based methods, such as 
\newcite{Goodfellow+:2014} and \newcite{carlini2017towards} for 
white-box attacks and \newcite{chen2017zoo} for black-box attacks.  For 
texts, we also have white-box attacks \cite{ebrahimi2017hotflip} and 
black-box attacks \cite{gao2018black,li2018textbugger} proposed in this 
category. 

Model-based attacks are generally seen as grey-box attacks as it 
requires access to target classifier during training phase, but once 
it's trained it can generate adversarial examples independently.  
\newcite{xiao2018generating} introduce a generative adversarial network 
to generate the image perturbation from a noise map. Generally in 
model-based attacks the attacking method and target classifier form a 
large network and the attacking method is trained using the loss from 
the target classifier. Note, however, that it is not very 
straightforward to use these model-based techniques for text directly  
because words in the adversarial examples are discrete and the network 
is not fully differentiable.


\section{Methodology}

\subsection{Sentiment Classifiers}
\label{sec:target-classifiers}

There are a number of off-the-shelf neural models for sentiment
classification \cite{kim2014convolutional,wang2016attention}, most
of which are based on long-short term memory
networks (LSTM; \newcite{hochreiter1997long}) or convolutional neural
networks (CNN; \newcite{kim2014convolutional}). In this paper,
we pre-train three sentiment classifiers: \mdbi,
\mdattn, and \mdcnn. These classifiers are targeted by different
attacking methods to generate adversarial examples (detailed in
\secref{attacking-methods}).  \mdbi is composed
of an embedding layer that maps individual words to pre-trained word
embeddings; a number of bi-directional LSTMs
that capture sequential contexts; and an output layer that maps the
averaged LSTM hidden states to a binary output.
\mdattn is similar to \mdbi except it has an extra
self-attention layer which learns to attend to salient words for
sentiment classification, and we compute the weighted mean of the LSTM
hidden states prior to the output layer.
Manual inspection of the attention weights show that polarity words such
as \textit{awesome} and \textit{disappointed} are assigned with higher
weights. Finally, \mdcnn has a number of convolutional filters of
varying sizes, and their outputs are concatenated, pooled and fed to
a fully-connected layer followed by a binary output layer.

Recent development in transformer-based pre-trained models have produced
state-of-the-art performance on a range of NLP tasks
\cite{devlin2018bert,Yang+:2019}.  To validate the transferability of the
attacking methods, we also test it on a fine-tuned \mdbert classifier.  
That is, we use the adversarial examples generated for
attacking the three previous classifiers (\mdbi, \mdattn or \mdcnn) as
test data for \mdbert and measure its classification performance to 
understand whether these adversarial examples can fool \mdbert.

\subsection{Benchmark Attacking Methods}
\label{sec:attacking-methods}

We experiment with six benchmark attacking methods for texts, ranging
from white-box attacks:
\fgm, \fgvm, \deepfool \cite{gong2018adversarial}, \hotflip 
\cite{ebrahimi2017hotflip}), and \tsai \cite{tsai2019adversarial}
to black-box attacks: \textfooler \cite{jin2019bert}. 

To perturb the discrete inputs, both \fgm and \fgvm introduce noises in
the word embedding space via the fast gradient method \cite{Goodfellow+:2014} and
reconstruct the input by mapping perturbed word embeddings to valid
words via nearest neighbour search. Between \fgm and \fgvm, the former
introduce noises that is proportional to the sign of the gradients while
the latter introduce perturbations proportional to the gradients 
directly.
The proportion is known as the \textit{overshoot} value and
denoted by $\epsilon$.  \deepfool uses the same trick to deal with
discrete inputs except that, instead of using the
fast gradient method, it uses the method introduced in 
\newcite{moosavi2016deepfool} for image to
search for an optimal direction to perturb the word embeddings.

Unlike the previous three methods, \hotflip and \tsai rely on performing one
or more atomic \textit{flip} operations to replace words while
monitoring the label change given by the target classifier. In \hotflip,
the directional derivatives w.r.t.\  {flip} operations are
calculated and the {flip} operation that results in the largest
increase in loss is selected.\footnote{While the original paper
explores both character flips and word flips, we test only word flips here.
The rationale is that introducing character flips to word-based
target classifiers is essentially changing word tokens to [unk],
which creates a confound for our experiments.}
\tsai is similar to \fgm, \fgvm and \deepfool in that it also uses nearest
neighbour search to map the perturbed embeddings to valid words, but 
instead of using the perturbed tokens directly, it uses greedy search or 
beam search to flip original tokens to perturbed ones one-at-a-time in 
order of their vulnerability.

\textfooler, the only black-box attack method tested, is a query-based 
method.  Since it assumes no access to the full architecture of the 
target classifier, it learns the order of vulnerability of tokens in an 
input sentence according to the change of prediction scores produced by 
the target classifier when a specific token is discarded. Once the order 
of vulnerability of words is identified, similar to \hotflip and \tsai, 
it greedily replaces tokens in the order of vulnerability one-at-a-time 
until the prediction of the target classifier changes. To ensure 
similarity between the adversarial examples and the original ones, the 
substituted tokens are selected to satisfy semantic, part-of-speech and 
sentence embedding similarity constraints.  Note that \textfooler uses  
the classifier's prediction scores to learn token vulnerability; in a 
more realistic black-box scenario the classifier may only reveal the 
predicted labels (without showing the underlying scores associated with 
each label).

\section{Experiments}
\subsection{Datasets}

We construct three datasets based on the Yelp reviews\footnote{\url{https://www.yelp.com/dataset}}
and the sentence-level Rotten Tomato (RT) movie reviews\footnote{\url{http://www.cs.cornell.edu/people/pabo/movie-review-data/}}. 
For Yelp, we binarise the ratings\footnote{Ratings$\geq$4
 is set as positive and ratings$\leq$2 as negative.}, and create
2 datasets, where we keep only reviews with $\leq$
 50 tokens (\dsyelpsmall) and $\leq$200 tokens (\dsyelplarge).
For RT (\dsrt), we directly use the binarised dataset which contains 
5331 positive and 5331 negative tokenized sentences. 
 We randomly partition both datasets  into train/dev/test sets (90/5/5 
 for \dsyelpsmall; 99/0.5/0.5 for \dsyelplarge; and 80/10/10 for \dsrt).  
For \dsyelpsmall and \dsyelplarge, we use spaCy\footnote{\url{https://spacy.io}} 
for tokenisation. 
We train and tune target classifiers (see \secref{target-classifiers}) 
using the training and development sets, and evaluate their performance 
on the original examples in the test sets as well as the adversarial examples 
 generated by attacking methods for the test sets. 
These datasets present a variation in the text length (e.g.\ the average
number of words for \dsrt, \dsyelpsmall and \dsyelplarge is 22, 34 and 82 words 
respectively) and training data size (e.g.\ \dsrt: 8K examples, \dsyelpsmall: 407K examples, 
and \dsyelplarge: 2M examples). 



\subsection{Implementation Details}

We use the pre-trained \dsglove embeddings \cite{pennington2014glove}
for the first 5 attacking methods and the counter-fitted word embedding
 \cite{mrkvsic2016counter} for \textfooler. For \fgm, \fgvm and \deepfool, we
tune $\epsilon$, the overshoot hyper-parameter (\secref{attacking-methods})
and keep the iterative step $n$
static (5).\footnote{We search for the best $\epsilon$ within a large 
range
(orders of magnitude in difference) to achieve a particular attacking
performance.}  For \tsai, besides $\epsilon$ we also tune the upper
limit of flipped words, ranging from 10\%--100\% of the maximum
length.  For \hotflip and \textfooler, we tune only the upper limit of 
flipped words, in the range of $[1, 7]$. 


For target classifiers, we tune batch size, learning rate, number of 
layers, number of units, attention size (for \mdattn), filter sizes and 
dropout probability (for \mdcnn). For \mdbert,
we use the default fine-tuning hyper-parameter values except for batch
size, where we adjust based on memory consumption. Note that after the target
classifiers are trained their weights are not updated when running the 
attacking methods.


\section{Evaluation}

We propose both automatic metrics and human evaluation strategies to
assess the quality of adversarial examples, based on four criteria defined
in \secref{intro}:
(a) attacking performance (i.e.\ how well they fool the classifier); (b)
textual and semantic similarity between the original input and the adversarial input;
(c) fluency of the adversarial example; and (d) label preservation.
Note that the automatic metrics only address the first 3 criteria (a, b and
c); we contend that criterion (d) requires manual evaluation, as the
judgement of whether the original label is preserved is inherently a
human decision.


\subsection{Automatic Evaluation: Metrics}

As sentiment classification is our target task, we use the standard
classification accuracy (\accuracy, the lower the better) to evaluate 
the attacking performance of adversarial examples (criterion (a)).

To assess the similarity between the original and (transformed)
adversarial examples (criterion (b)), we compute \bleu scores 
\cite{Papineni+:2002} to measure word overlap; and \use scores, cosine 
similarity between the representations of original examples and 
adversarial examples generated by the universal sentence encoder 
\cite{cer2018universal}, to measure semantic  similarity.  For both  
metrics, higher
scores represent better performance. 

To measure fluency, we first explore a
supervised BERT model fine-tuned to predict linguistic acceptability
\cite{devlin2018bert}. However, in preliminary experiments we found that
BERT performs very poorly at predicting the acceptability of
adversarial examples (e.g.\ it predicts word-salad-like sentences
generated by \fgvm as very acceptable),
revealing the brittleness of these supervised models. We next explore 
unsupervised approaches \cite{Lau+:2017,Lau+:2020},  using normalised
sentence probabilities estimated by pre-trained language models for
measuring acceptability. Following \newcite{Lau+:2020},  we use XLNet
\cite{Yang+:2019} as the language model. The acceptability score of a 
sentence is calculated based on the normalised sentence probability: 
${\log P(s)} /
({((5+|s|)/(5+1))^\alpha})$, where $s$ is the sentence, and $\alpha$ is
a hyper-parameter (set to
0.8) to dampen the impact of large values \cite{Vaswani+:2017}.  To 
measure how fluency differs between an adversarial example and the 
original sentence, we compute the difference in their acceptability 
scores, giving us the acceptability metric \acceptability.

Note that we only compute \bleu and \acceptability for adversarial 
examples that have successfully fooled the classifier.  Our rationale is 
that unsuccessful examples can artificially boost these scores by not 
making any modifications, and so the better approach is to only consider 
successful examples.

\subsection{Automatic Evaluation: Results}

We present the performance of the attacking methods against 3 target
classifiers (Table \ref{tab:rc-full}A; top) and on 3 datasets (Table
\ref{tab:rc-full}B; bottom). We choose 3 \accuracy thresholds for the 
attacking performance: T0, T1 and T2, which correspond approximately to 
accuracy scores of 90\%, 80\% and 70\% for the Yelp datasets 
(\dsyelpsmall, \dsyelplarge)\footnote{Most 
methods are capable of achieving attacking performance of 30\% or lower 
(\accuracy), although that comes with severe degradation to the quality 
of the adversarial examples.}; and 60\%, 50\%, 30\% for RT dataset (\dsrt)
\footnote{We choose 30\% instead of 40\% because, for \hotflip, 
flipping one words achieved an accuracy drop to 30.1\%}.
Our rationale is that each method should 
be compared on the same basis  if our focus is to to provide a fair 
assessment on the quality of the adversarial examples, and the attacking 
performance constitutes a reasonable basis.  


Looking at \bleu, \use and \acceptability, \textfooler is the most consistent 
method over multiple datasets and classifiers. \hotflip is also fairly 
competitive, occasionally producing better \bleu scores (\mdcnn at T1; 
at T0, T1, T2 on \dsyelplarge and T2 on \dsrt).
Gradient-based methods \fgm and \fgvm perform very poorly. In general, 
they tend to produce word salad adversarial examples, as indicated by 
their poor \acceptability scores. \deepfool similarly generates incoherent 
sentences with low \bleu scores, but occasionally produces good \use 
(at T0) and \acceptability (\mdbi at T1 and T2), suggesting potential brittleness 
of the automatic evaluation approach for evaluating semantic similarity and
acceptability.

\begin{table*}[t!]
\begin{center}
\begin{adjustbox}{max width=0.6\linewidth}
\begin{tabular}{ll|r@{\;}r@{\;}r@{\;}r|r@{\;}r@{\;}r@{\;}r|r@{\;}r@{\;}r@{\;}r}
\toprule
 \multicolumn{2}{c}{(A)}  & \multicolumn{12}{c}{Dataset: \dsyelpsmall} \\
 \midrule
  \multicolumn{2}{c}{Models: acc} & \multicolumn{4}{c}{\mdattn: 96.8}  & 
 \multicolumn{4}{c}{\mdcnn: 94.3} & \multicolumn{4}{c}{\mdbi: 96.6}\\
 \midrule
 \multicolumn{2}{c|}{Attack} & \accuracy & \bleu & \use & \acceptability & 
  \accuracy & \bleu & \use & \acceptability & \accuracy & \bleu  & \use & 
\acceptability\\
  \midrule
\multirow{6}{*}{T0} & \fgm & 93.0 & 34.6 & 17.4 & -25.9 & 92.3 & 65.9 & 47.1 & -16.6 & 
 90.5 & 30.0 & 2.3 & -25.9\\
 & \fgvm & 93.4 & 29.1 & 10.1 & -17.7 & 93.4 & \textbf{89.0} & \textbf{80.0} & \textbf{-5.8} & 92.9 & 19.6 & 16.2 & -21.4\\
 & \deepfool & 94.7 & 20.1 & 61.7 & -20.6 & 92.7 & 68.6 & 72.8 & -15.0 & 93.9 & 16.0 & \textbf{68.8} & -17.8\\
 & \tsai & 91.7 & 64.7 & 38.9 & -14.4 & 90.4 & 59.0 & 41.5 & -16.9 & 90.8 & 65.4 & 34.9 & -13.6\\
 & \hotflip & 92.5 & \textbf{92.6} & \textbf{66.5} & \textbf{-3.7} & -- & -- & -- & -- & 93.2 & \textbf{92.7} & 67.3 & \textbf{-3.5} \\
  & \textfooler & -- & -- & -- & -- & -- & -- & -- & -- & -- & -- & -- & -- \\
 \midrule
\multirow{6}{*}{T1} & \fgm & 88.7 & 15.4 & -13.1 & -28.9 & 82.2 & 16.3 & -7.1 & -35.2 & 
 81.0 & 11.6 & -15.6 & -27.7\\
 & \fgvm & 83.9 & 7.6  & -24.2 & -12.1 & 82.6 & 20.6 & 2.2 & -34.0 & 85.4 & 11.4 & -13.9 & -15.8\\
 & \deepfool & 86.8 & 13.4 & 27.6 & -10.1 & 84.6 & 17.5 & 33.4 & -34.4 & 80.8 & 5.6 & 14.1 & \textbf{-0.7} \\
 & \tsai & 83.8 & 48.3 &  11.6 & -18.9 & 87.6 & 41.2 & 29.4 & -21.8 & 81.8 & 47.4 & 8.9 & -19.0\\
 & \hotflip & 80.3 & 85.6 & 47.9 & -7.0 & 81.5 & \textbf{92.5} & 77.1 & \textbf{-3.8} & 82.8 & 85.1 & 47.6 & -7.0\\
   & \textfooler & 86.5 & \textbf{92.6} & \textbf{88.7} & \textbf{-1.8} & 87.7 & 91.9 & \textbf{94.2} & -2.1 & 85.8 & \textbf{92.8} & \textbf{82.1} & -1.6 \\
 \midrule
\multirow{6}{*}{T2} & \fgm & 72.7 & 2.7  & -33.6 & -34.5 & 71.9 & 2.4 & -30.8 & -38.5 & 
 71.4 & 3.3 & -26.3 & -28.8\\
 & \fgvm & 77.8 & 4.6 & -20.9 & -9.3 & 71.7 & 7.0 & -18.7 & -38.3 & 70.9 & 0.3 & -37.0 & -4.6\\
 & \deepfool & 72.1 & 3.1 & -28.5 & -12.6 & 70.9 & 5.4 & -20.8 & -38.3 & 72.0 & 2.9 & 6.0 & \textbf{0.5} \\
 & \tsai & 75.3 & 41.2 & -7.6 & -20.7 & 73.4 & 38.9 & -15.3 & -21.4 & 77.5 & 43.1 & 0.6 & -19.9\\
 & \hotflip & 75.3 & 80.0 & 38.1  & -7.8 & 70.8 & \textbf{84.7} & \textbf{63.4} & \textbf{-7.1} & 70.6 & 78.7 & 36.7 & -9.8\\
    & \textfooler & 73.6 & \textbf{88.5} & \textbf{84.1} & \textbf{-2.9} & -- & -- & -- & -- & 70.8 & \textbf{88.4} & \textbf{86.9} & -2.77 \\
\bottomrule
\toprule
 \multicolumn{2}{c}{(B)} & \multicolumn{12}{c}{Target classifier: 
 \mdattn} \\
 \midrule
 \multicolumn{2}{c}{Datasets: acc} & \multicolumn{4}{c}{\dsyelpsmall: 
   96.8}  & \multicolumn{4}{c}{\dsyelplarge: 97.9} & \multicolumn{4}{c}{\dsrt: 78.8 }\\
   \midrule
  \multicolumn{2}{c|}{Attack} & \accuracy & \bleu & \use & \acceptability & 
  \accuracy & \bleu & \use & \acceptability  & \accuracy & \bleu  & \use & 
\acceptability \\
  \midrule
\multirow{6}{*}{T0} & \fgm & 93.0 & 34.6 & 17.4 & -25.9  & 92.1 & 13.8 & 3.2 & -37.8  & 66.3 & 4.9 & -15.9 & -29.1 \\
 & \fgvm & 93.4 & 29.1 & 10.1 & -17.7 & 94.2 & 55.4 & 63.7 & -18.3 & 66.3 & 25.3 & 13.3 & -24.4 \\
 & \deepfool & 94.7 & 20.1 & 61.7 & -20.6 & -- & -- & -- & -- & 62.2 & 2.1 & -31.9 & -5.9  \\
 & \tsai & 91.7 & 64.7 & 38.9 & -14.4 & 90.3 & 51.2 & 44.5 & -20.3  & 65.4 & \textbf{67.2} & \textbf{33.2} & \textbf{-9.3} \\
 & \hotflip & 92.5 & \textbf{92.6} & \textbf{66.5} & \textbf{-3.7} & 90.8 & \textbf{96.4} & 75.6 & -3.3 & -- & -- & -- & --  \\
 & \textfooler & -- & -- & -- & -- & 93.8 & 96.3 & \textbf{94.4} & \textbf{-1.6} & -- & -- & -- & --  \\
 \midrule
\multirow{6}{*}{T1}& \fgm & 88.7 & 15.4 & -13.4 & -28.9 & 81.6 & 23.0 & 6e-3 & -37.0 & -- & -- & -- & -- \\
 & \fgvm & 83.9 & 7.6 & -24.2 & -12.1 & 80.8 & 17.2 & 10.6 & -35.7 & 48.8 & 7.1 & -11.6 & -23.7 \\
 & \deepfool & 86.8 & 13.4 & 27.6 & -10.1 & 82.3 & 19.1 & -0.4 & -9.0 & 49.8 & 0.4 & -41.7 & -18.6 \\
 & \tsai & 83.8 & 48.3 & 11.6 & -18.9 & 86.9 & 42.6 & 35.7 & -23.7 &  50.6 & 46.0 & 18.7 & -16.1  \\
 & \hotflip & 80.3 & 85.6 & 47.9 & -7.0 & 83.2 & \textbf{94.8} & 67.2 & -4.1 & -- & -- & -- & --  \\
  & \textfooler & 86.5 & \textbf{92.6} & \textbf{88.7} & \textbf{-1.8} & 84.2 & 92.6 & \textbf{90.3} & \textbf{-3.3} & 49.9 & \textbf{87.7} & \textbf{73.0} & \textbf{-3.5}  \\
  \midrule
\multirow{6}{*}{T2} & \fgm & 72.7 & 2.7 & -33.6 & -34.5 & 72.8 & 6.9 & 14.8 & -28.5 & -- & -- & -- & -- \\
 & \fgvm & 77.8 & 4.6 & -20.9 & -9.3 & 70.6 & 1.5 & -26.5 & -5.4 & -- & -- & -- & -- \\
 & \deepfool & 72.1 & 3.1 & -28.5 & -12.6 & 72.2 & 7.7 & 25.2 & -25.4 & -- & -- & -- & -- \\
 & \tsai & 75.3 & 41.2 & -7.6 & -20.7 & 76.9 & 36.3 & 8.7 & -23.5 & -- & -- & -- & -- \\
 & \hotflip & 75.3 & 80.0 & 38.1 & -7.8 & 77.0 & \textbf{93.8} & 62.9 & -4.7 & 30.1 & \textbf{89.0} & 64.5 & \textbf{-5.0}  \\
   & \textfooler & 73.6 & \textbf{88.5} & \textbf{84.1} & \textbf{-2.9} & 75.7 & 90.2 & \textbf{88.2} & \textbf{-4.5} & 33.7 & 82.0 & \textbf{69.1} & -5.5  \\
\bottomrule
\end{tabular}
\end{adjustbox}
\end{center}
\caption{Results based on automatic metrics. Top half (A) presents 3
different target classifiers evaluated on one dataset (\dsyelpsmall); 
bottom half (B) tests 3 datasets using one classifier (\mdattn). For 
\acceptability, less negative values are better. Boldface indicates 
optimal performance for an attacking performance and target classifier.  
Missing numbers (dashed lines) indicate the method is unable to produce 
the desired accuracy, e.g.\ \hotflip with only 1 word flip produces  
81.5\% accuracy (T1) when attacking \mdcnn on \dsyelpsmall, and so T0 
accuracy is unachievable.}
\label{tab:rc-full}
\end{table*}

Comparing the performance across different \accuracy thresholds, we 
observe a consistent pattern of decreasing performance over all metrics 
as the attacking performance increases from T0 to T2, especially drastic 
for \fgm, \fgvm and \deepfool when the attacking rate changes from T0 to 
T1.  These 
observations suggest that all methods are trading off fluency and 
content preservation as they attempt to generate stronger adversarial 
examples.


We now focus on \tabref{rc-full}A to understand the impact of model 
 architecture for the target classifier.  With 1 word flip as the upper limit 
 for \hotflip, the accuracy of \mdattn and \mdbi drops to T0 (approximately 
 4\% accuracy decrease) while the accuracy of \mdcnn drops to T1 
 (approximately 13\% accuracy decrease). 
Within the same attacking performance thresholds, \fgm, \fgvm and \deepfool achieve
higher \bleu, \use and \acceptability scores when targeting \mdcnn compared to 
those scores when targeting the other two models. 
 These observations suggest that convolutional 
 networks are more vulnerable to attacks (noting that it is the predominant 
 architecture for CV).  Interestingly, the \mdcnn model seems to be very robust
 against \textfooler on \dsyelpsmall, where the accuracy stays at around  
87.0\% regardless of how we tune \textfooler. 
 
Looking at \tabref{rc-full}B, we also find that the attacking performance is 
influenced by the input text length and the number of training examples for target
classifiers. For \hotflip and \textfooler, we see improvements over \bleu, \use and 
\acceptability as text length increases from \dsyelpsmall to \dsyelplarge, indicating
the performance of these two methods is more affected by input lengths. We think 
this is because with more words it is more likely for them to find a vulnerable spot 
to target. 
While for \tsai, we see improvements over \bleu and \acceptability 
as the number of training examples for target classifier decreases from 
\dsyelplarge (2M) to \dsyelpsmall (407K), indicating \tsai are 
less effective for attacking target classifiers that are trained with 
more data. This suggests that increasing the training data for a 
classifier could potentially improve its robustness against certain 
attacks. 
The effect of the number of training examples for target classifier is further
validated by the attacking performance of \textfooler and \hotflip on \dsrt.
Despite that \dsrt has the shortest input (with average 22 tokens), changing
1 words for \textfooler and \hotflip successfully rendered accuracy drops from
78.8\% to 49.9\% and 30.1\%, respectively. Comparing the same number of 
word change, the drop of accuracy introduced by \textfooler and \hotflip are 10\% 
and 4\% on \dsyelpsmall; and 4\% and 7\% on \dsyelplarge. 

Another factor that possibly makes it easier to attack \dsrt is that movie 
reviews being more descriptive and therefore creating potential ambiguity 
in their expression of sentiment.
 In comparison, the restaurant reviews are more straightforward,
using polarising words such as \textit{awesome} or \textit{awful}.
The net effect is that sentiment classification is ``easier'' for the 
restaurant reviews (as the classifier can make the decision based on 
vocabulary choices), which in turn make adversarial attacks ``harder''.  



\begin{figure}[!t]
  \centering
  \includegraphics[width=0.48\textwidth]{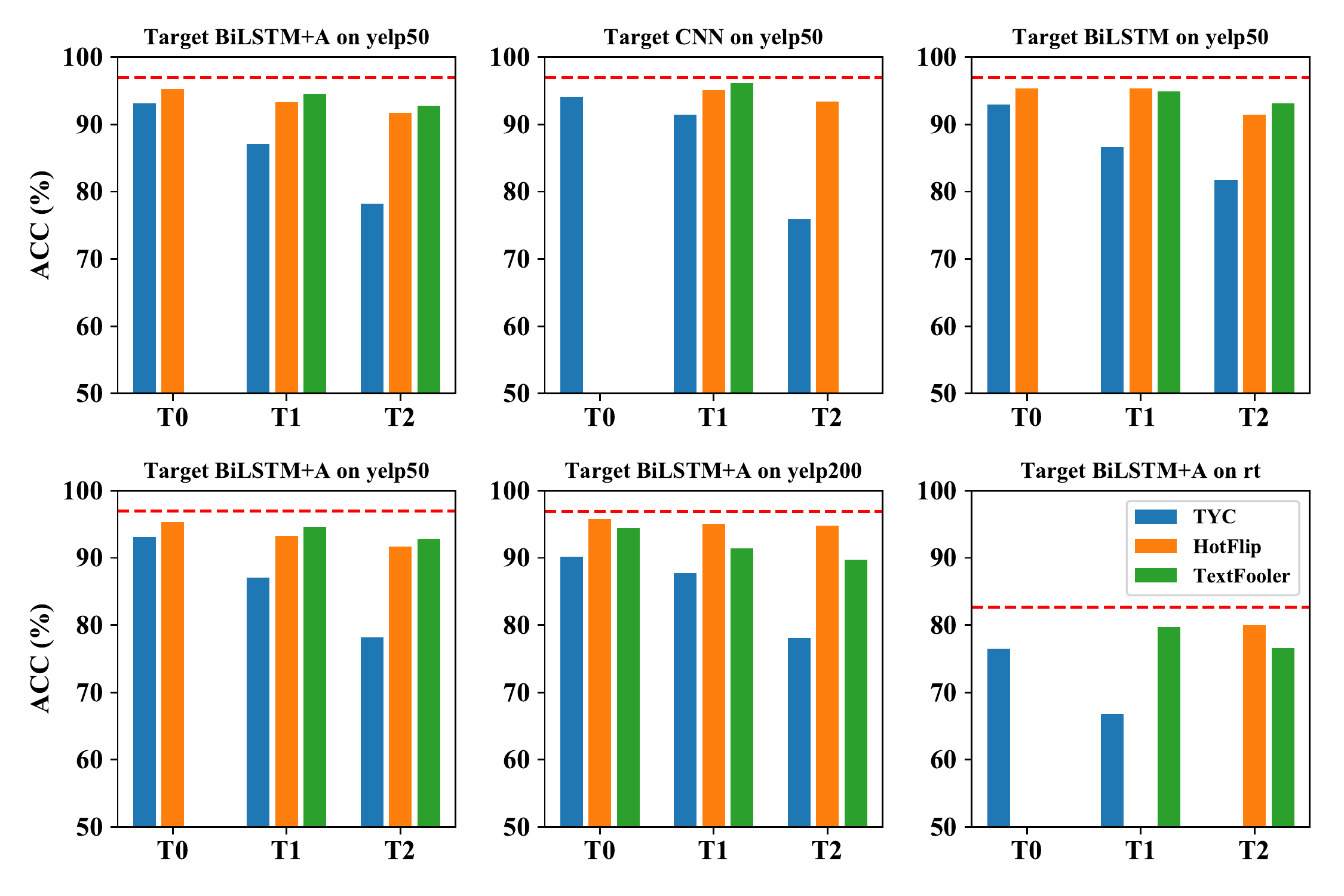}
  \caption{Accuracy of \mdbert on adversarial examples generated from 
  \tsai, \hotflip and \textfooler for different target classifiers (top row) and 
  for different datasets (bottom row). 
  \mdbert's accuracy on original examples are denoted as red line in each 
  plot.}
  \label{fig:figure2}
\end{figure}

To check how well these adversarial examples generalise to fooling other 
classifiers, also known as \textit{transferability},  we feed the adversarial 
examples from the 3 best methods, i.e. \tsai, \hotflip and \textfooler, 
to a pre-trained \mdbert trained for sentiment classification  and 
measure its accuracy (\figref{figure2}).  Unsurprisingly, we observe 
that the attacking performance (i.e.\ the drop in \accuracy) is not as 
good as those reported in \tabref{rc-full} .  Interestingly, we find 
that \textfooler, the best performing method, produces the least effective 
adversarial examples for \mdbert, indicating poor transferability, 
while examples generated from \tsai perform the best in fooling \mdbert. 
Manually inspecting the examples generated by 
\textfooler, we notice it tries to replace as few words as possible to 
move an examples just across the decision boundary of the target
classifier (indicated by very close scores between different labels). 
This change appears be very targeted for a specific classifier, and as 
such is unlikely to fool other classifiers.

As an additional insight, we also evaluated the computational time of different
attacking methods. We calculate the the generation time of the three 
best performing
methods \tsai, \hotflip and \textfooler when they attack the test dataset of
\dsrt, \dsyelpsmall, and \dsyelplarge at T2 attacking threshold.  \tabref{rc-time} 
shows the corresponding computational time (seconds per example). We 
found  that \hotflip achieves comparable performance with \textfooler 
but only consumes 1/3 of its computational time. \tsai is even more time-consuming 
than \textfooler. Also, the speed of \hotflip seems not affected by the increase
of the sentence lengths, but \textfooler and \tsai take much longer time 
to attack longer input examples. 

To summarise, our results demonstrate that the best attacking methods 
(e.g.\ \textfooler and \hotflip) may not produce adversarial examples that 
generalise to fooling other classifiers. We also saw that convolutional networks 
are generally more vulnerable than recurrent networks, and that dataset 
features such as text length and training data size can influence the 
performance of adversarial attacks.


\subsection{Human Evaluation: Design}

Automatic metrics provide a proxy to quantify the quality of the 
adversarial examples. To validate that these metrics work, 
we conduct a crowdsourcing experiment on 
Figure-Eight.\footnote{\url{https://www.figure-eight.com/}} Recall that 
the automatic metrics do not assess sentiment preservation (criterion 
(d)); we evaluate that aspect here.

\begin{table}[t]
\scriptsize
\begin{center}
\begin{tabular}{lrrr}
\toprule
 Methods & \dsrt  & \dsyelpsmall & \dsyelplarge \\
 \midrule
\tsai & -- & 1.2 & 13.8  \\
\hotflip & 0.06 & 0.3 & 0.6  \\
\textfooler & 0.5 & 1.0 & 8.0  \\
\bottomrule
\end{tabular}
\end{center}
\caption{Computational time (seconds per example) of \tsai, \hotflip and 
\textfooler when attacking different datasets. }
\label{tab:rc-time}
\end{table}


We experiment with the 3 best methods (\tsai \hotflip and \textfooler) on 2 
accuracy thresholds (T0 and T2), using \mdattn as the classifier. For 
each method and threshold, we randomly sample 25 positive-to-negative 
and 25 negative-to-positive examples. To control for quality, we reserve 
and annotate 10\% of the samples ourselves as control questions. Workers 
are first presented with 10 control questions as a quiz, and only those 
who pass the quiz with at least 80\% accuracy can continue to work on 
the task.  We display 10 questions per page, where one control question 
is embedded to continue monitor worker performance.\footnote{Workers are 
required to maintain their performance (80\% accuracy) throughout the 
annotation process.}  The task is designed such that each control 
question can only be seen once per worker. We restrict our jobs to 
workers in United States, United Kingdoms, Australia, and Canada.

To evaluate the criteria (b) textual similarity, (c) fluency, and (d) 
sentiment preservation, we ask the annotators three questions:
\begin{enumerate}
\setlength\itemsep{0em}
  \item Is snippet B a good paraphrase of snippet A? \newline
  \hspace{4pt} $\ocircle$ Yes \hspace{4pt} $\ocircle$ Somewhat yes \hspace{4pt} $\ocircle$ No
  \item How natural does the text read? \newline
  \hspace{4pt} $\ocircle$ Very unnatural \hspace{4pt} $\ocircle$ Somewhat natural \hspace{4pt} $\ocircle$ Natural
   \item What is the sentiment of the text? \newline
  \hspace{4pt} $\ocircle$ Positive \hspace{4pt} $\ocircle$ Negative \hspace{4pt} $\ocircle$ Cannot tell
\end{enumerate}

For question 1, we display both the adversarial input and the original 
input, while for question 2 and 3 we present only the adversarial 
example.  As a baseline, we also run a survey on question 2 and 3 for 
50 random original (unperturbed) samples.


\begin{figure}[!t]
  \begin{minipage}[b]{0.48\textwidth}%
    \centering
    \subfloat[Original 
  examples]{\includegraphics[width=\linewidth]{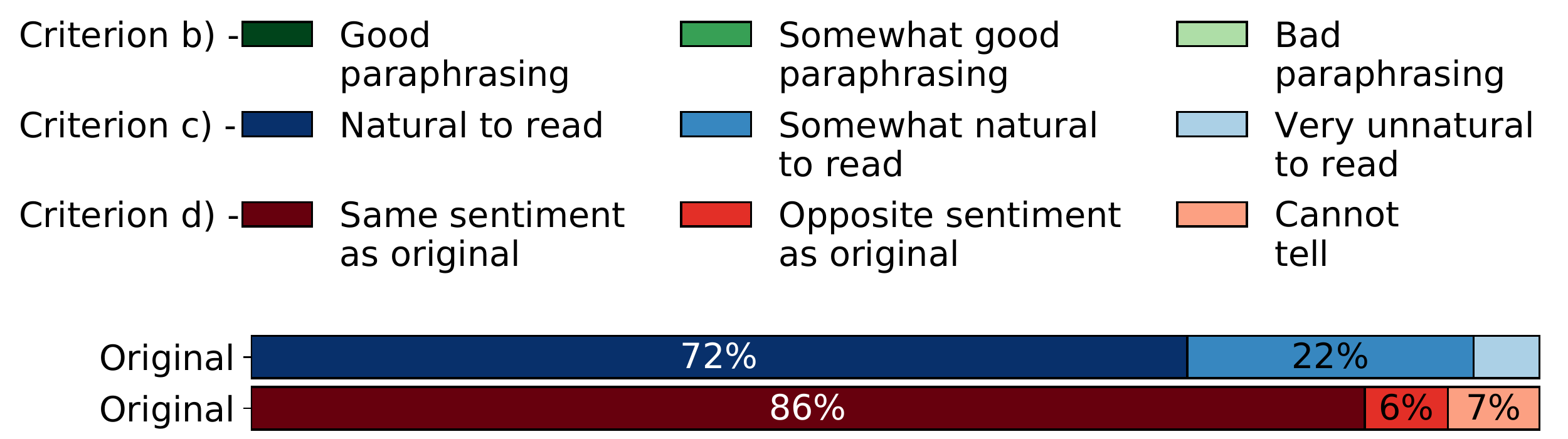}}%
  \end{minipage}\\
  \hfill
  \begin{minipage}[b]{0.48\textwidth}%
    \centering
    \subfloat[\accuracy threshold:  
  T0]{\includegraphics[width=\textwidth]{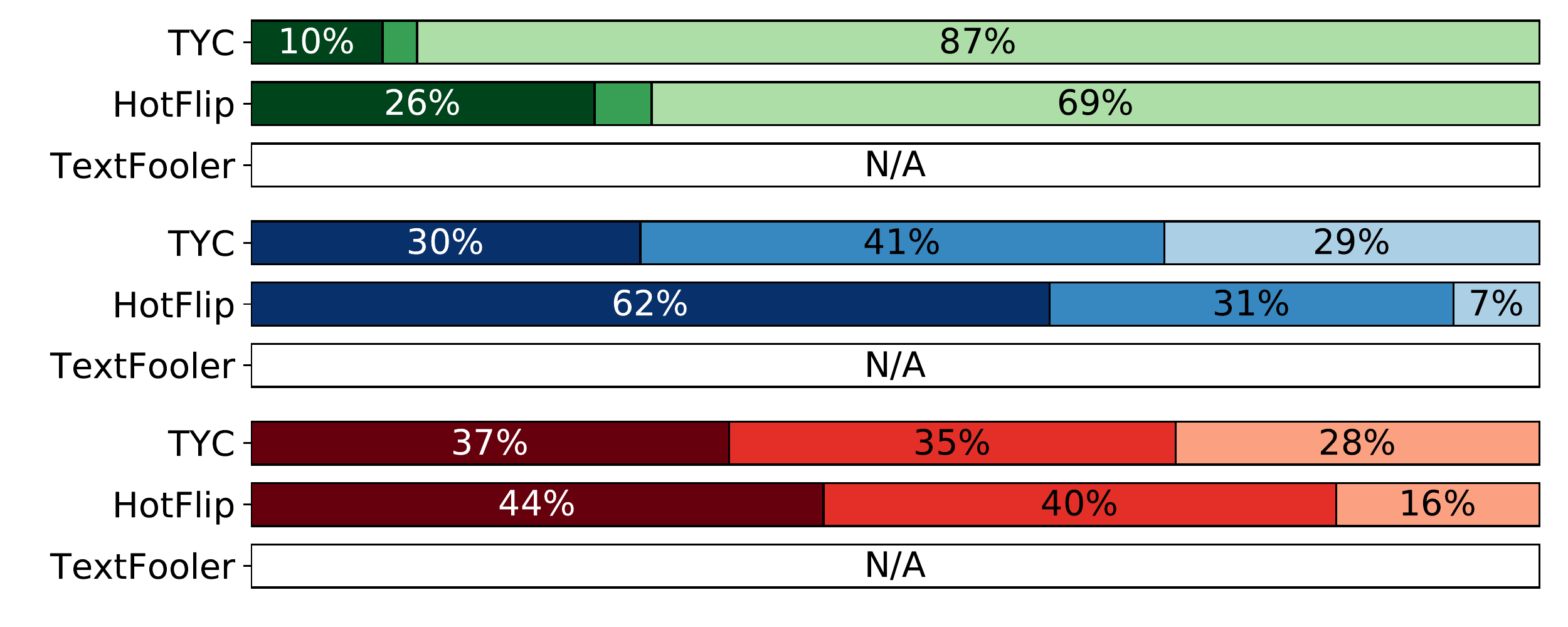}}%
  \end{minipage}\\
  \hfill
  \begin{minipage}[b]{0.48\textwidth}%
    \centering
    \subfloat[\accuracy threshold: 
  T2]{\includegraphics[width=\textwidth]{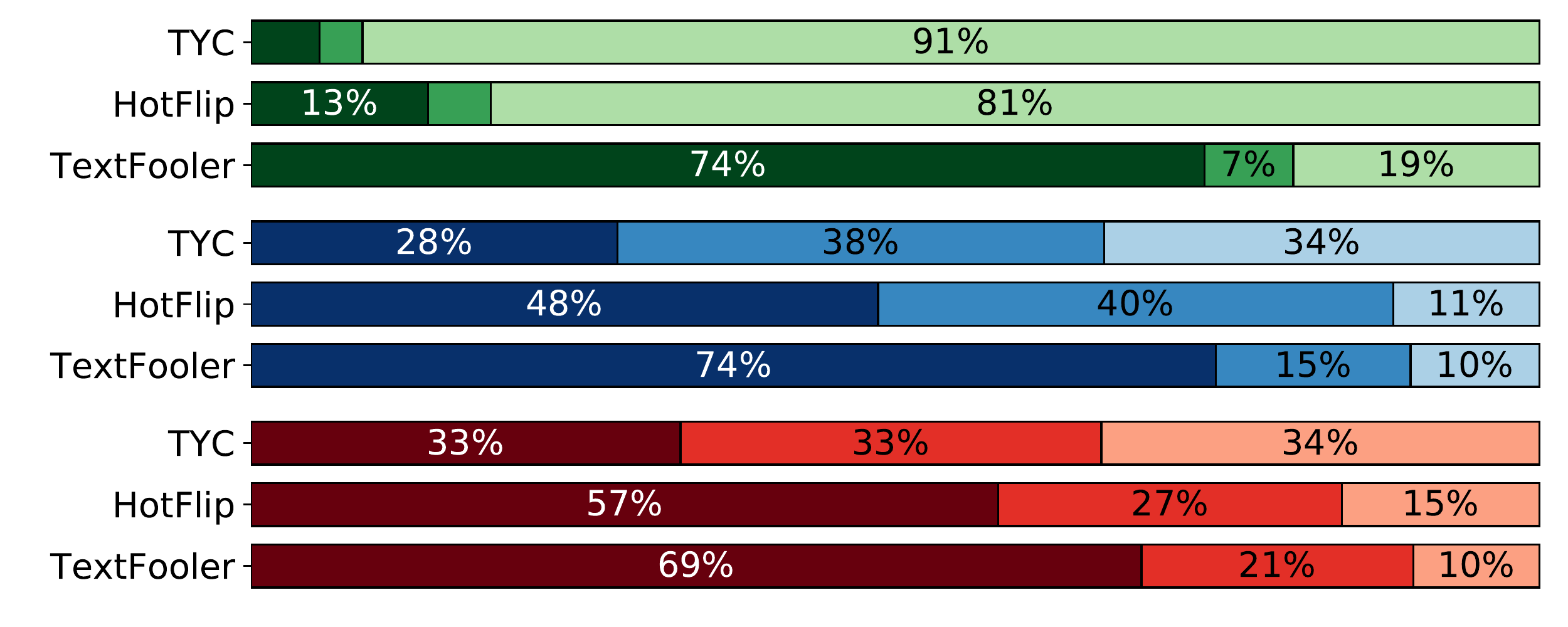}}%
  \end{minipage}
  \caption{Human evaluation results.}
  \label{fig:human_eval}
\end{figure}

\subsection{Human Evaluation: Results}

We present the percentage of answers to each question  in 
\figref{human_eval}. The green bars illustrate how well the adversarial 
examples paraphrase the original ones; blue how natural the 
adversarial examples read; and red whether the sentiment of the 
adversarial examples is consistent compared to the original.

Looking at the performance of the original sentences (``(a) Original 
samples''), we see that their language is largely fluent and their  
sentiment is generally consistent to the original examples'. 



On content preservation (criterion (b); green bars), all methods produce 
poor paraphrases on \dsyelpsmall except for \textfooler. 

Next we look at fluency (criterion (c); blue bars). 
We see similar trend: with the increased attacking performance, the 
readability of adversarial examples generated by different attacking 
methods is getting poorer. 
\hotflip is fairly competitive, producing adversarial examples that are only 
marginally less fluent compared to the original at T0. At T2, however, 
it begin to trade off fluency. 
\textfooler, on the other hand, achieved even slightly better fluency than the
original examples at T2, indicating the substituted tokens fit the 
context very well.  

Lastly, we consider sentiment preservation (criterion (d); red bars).
All methods trade off sentiment preservation to achieve better
attacking performance. 
Again both \hotflip and \textfooler are the better methods here 
(interestingly, we observe an increase in agreement 
as their attacking performance increases from T0 and T2).

Comparing automatic evaluation and human evaluation results, we found 
that the \use scores are generally consistent with the human evaluation 
results on semantic preservation, while the \bleu score is less 
effective as evidenced by the high \bleu score of \hotflip on 
\dsyelpsmall and poor paraphrasing performance in the same settings.  
The \acceptability metric appears to be a solid metric in evaluating 
fluency, as we see a good agreement with human evaluation across the 
three attacking methods. 


Summarising our findings, \textfooler is generally the best method across 
all criteria, noting that its adversarial examples, however, have the poorest 
transferability. \hotflip produces comparable results with \textfooler for 
meeting the four criteria and similarly suffer from poor transferability.  
\tsai generates adversarial examples with better transferability
but do not do well in terms of content preservation and fluency. 
In terms of computational time, \hotflip is the most efficient, consuming
less than 1/3 of the time consumed by the other two methods. The difference
is more profound for longer input sentences. 
\fgm, \fgvm and \deepfool perform very poor as they largely sacrifice example 
qualities to achieve attacking performance, indicating directly mapping from 
perturbed word embedding is not applicable in NLP. 
All said, we found that all 
methods tend to trade-off sentiment preservation for attacking performance, 
revealing that these methods in a way ``cheat'' by 
simply flipping the sentiments of the original sentences to fool the 
classifier, and therefore the adversarial examples might be ineffective 
for adversarial training, as they are not examples that reveal potential 
vulnerabilities in the classifier.

\section{Conclusion}

We propose an evaluation framework for assessing the quality of 
adversarial examples in NLP, based on four criteria: (a) attacking 
performance, (b) textual similarity; (c) fluency; (d) label 
preservation. Our framework involves both automatic and human 
evaluation, and we test 6 benchmark methods involving both white-box
and black-box attacking methods. 
We found that the architecture of the target classifier is an 
important factor when it comes to attacking performance, e.g.\ CNNs are 
more vulnerable than LSTMs. Data features such as length of text and 
input domains are also influencing factors that affect how difficulty it 
is to perform adversarial attack. 
Lastly, 
we observe in our human evaluation that on short texts that express 
clear positive or negative sentiments (such as \dsyelpsmall), these 
methods produce adversarial examples that tend not to preserve their 
semantic content and have low readability.  More importantly, these 
methods ``cheat'' by simply flipping the sentiment in the adversarial 
examples, and this behaviour is evident especially on the \dsyelpsmall 
dataset, suggesting they could be ineffective for adversarial training.


\bibliographystyle{acl_natbib}
\bibliography{emnlp2020}

\begin{thebibliography}{27}
\expandafter\ifx\csname natexlab\endcsname\relax\def\natexlab#1{#1}\fi

\bibitem[{Carlini and Wagner(2017)}]{carlini2017towards}
Nicholas Carlini and David Wagner. 2017.
\newblock Towards evaluating the robustness of neural networks.
\newblock In \emph{2017 IEEE Symposium on Security and Privacy (SP)}, pages
  39--57. IEEE.

\bibitem[{Cer et~al.(2018)Cer, Yang, yi~Kong, Hua, Limtiaco, John, Constant,
  Guajardo-Cespedes, Yuan, Tar, Sung, Strope, and Kurzweil}]{cer2018universal}
Daniel Cer, Yinfei Yang, Sheng yi~Kong, Nan Hua, Nicole Limtiaco, Rhomni~St.
  John, Noah Constant, Mario Guajardo-Cespedes, Steve Yuan, Chris Tar,
  Yun-Hsuan Sung, Brian Strope, and Ray Kurzweil. 2018.
\newblock \href {http://arxiv.org/abs/1803.11175} {Universal sentence encoder}.

\bibitem[{Chen et~al.(2017)Chen, Zhang, Sharma, Yi, and Hsieh}]{chen2017zoo}
Pin-Yu Chen, Huan Zhang, Yash Sharma, Jinfeng Yi, and Cho-Jui Hsieh. 2017.
\newblock Zoo: Zeroth order optimization based black-box attacks to deep neural
  networks without training substitute models.
\newblock In \emph{Proceedings of the 10th ACM Workshop on Artificial
  Intelligence and Security}, pages 15--26. ACM.

\bibitem[{Devlin et~al.(2018)Devlin, Chang, Lee, and
  Toutanova}]{devlin2018bert}
Jacob Devlin, Ming-Wei Chang, Kenton Lee, and Kristina Toutanova. 2018.
\newblock Bert: Pre-training of deep bidirectional transformers for language
  understanding.
\newblock \emph{arXiv preprint arXiv:1810.04805}.

\bibitem[{Dong et~al.(2018)Dong, Liao, Pang, Su, Zhu, Hu, and
  Li}]{dong2018boosting}
Yinpeng Dong, Fangzhou Liao, Tianyu Pang, Hang Su, Jun Zhu, Xiaolin Hu, and
  Jianguo Li. 2018.
\newblock Boosting adversarial attacks with momentum.
\newblock In \emph{Proceedings of the IEEE conference on computer vision and
  pattern recognition}, pages 9185--9193.

\bibitem[{Ebrahimi et~al.(2017)Ebrahimi, Rao, Lowd, and
  Dou}]{ebrahimi2017hotflip}
Javid Ebrahimi, Anyi Rao, Daniel Lowd, and Dejing Dou. 2017.
\newblock Hotflip: White-box adversarial examples for text classification.
\newblock \emph{arXiv preprint arXiv:1712.06751}.

\bibitem[{Gao et~al.(2018)Gao, Lanchantin, Soffa, and Qi}]{gao2018black}
Ji~Gao, Jack Lanchantin, Mary~Lou Soffa, and Yanjun Qi. 2018.
\newblock Black-box generation of adversarial text sequences to evade deep
  learning classifiers.
\newblock In \emph{2018 IEEE Security and Privacy Workshops (SPW)}, pages
  50--56. IEEE.

\bibitem[{Gong et~al.(2018)Gong, Wang, Li, Song, and Ku}]{gong2018adversarial}
Zhitao Gong, Wenlu Wang, Bo~Li, Dawn Song, and Wei-Shinn Ku. 2018.
\newblock Adversarial texts with gradient methods.
\newblock \emph{arXiv preprint arXiv:1801.07175}.

\bibitem[{Goodfellow et~al.(2014)Goodfellow, Shlens, and
  Szegedy}]{Goodfellow+:2014}
Ian~J Goodfellow, Jonathon Shlens, and Christian Szegedy. 2014.
\newblock Explaining and harnessing adversarial examples.
\newblock \emph{arXiv preprint arXiv:1412.6572}.

\bibitem[{Hochreiter and Schmidhuber(1997)}]{hochreiter1997long}
Sepp Hochreiter and J{\"u}rgen Schmidhuber. 1997.
\newblock Long short-term memory.
\newblock \emph{Neural computation}, 9(8):1735--1780.

\bibitem[{Jin et~al.(2019)Jin, Jin, Zhou, and Szolovits}]{jin2019bert}
Di~Jin, Zhijing Jin, Joey~Tianyi Zhou, and Peter Szolovits. 2019.
\newblock \href {http://arxiv.org/abs/1907.11932} {Is bert really robust? a
  strong baseline for natural language attack on text classification and
  entailment}.

\bibitem[{Kim(2014)}]{kim2014convolutional}
Yoon Kim. 2014.
\newblock Convolutional neural networks for sentence classification.
\newblock \emph{arXiv preprint arXiv:1408.5882}.

\bibitem[{Kurakin et~al.(2016)Kurakin, Goodfellow, and
  Bengio}]{kurakin2016adversarial}
Alexey Kurakin, Ian Goodfellow, and Samy Bengio. 2016.
\newblock Adversarial examples in the physical world.
\newblock \emph{arXiv preprint arXiv:1607.02533}.

\bibitem[{{Lau} et~al.(2020){Lau}, {Armendariz}, {Lappin}, {Purver}, and
  {Shu}}]{Lau+:2020}
Jey~Han {Lau}, Carlos~S. {Armendariz}, Shalom {Lappin}, Matthew {Purver}, and
  Chang {Shu}. 2020.
\newblock \href {http://arxiv.org/abs/2004.00881} {How furiously can colourless
  green ideas sleep? sentence acceptability in context}.
\newblock \emph{arXiv e-prints}, page arXiv:2004.00881.

\bibitem[{Lau et~al.(2017)Lau, Clark, and Lappin}]{Lau+:2017}
Jey~Han Lau, Alexander Clark, and Shalom Lappin. 2017.
\newblock Grammaticality, {Acceptability}, and {Probability}: {A}
  {Probabilistic} {View} of {Linguistic} {Knowledge}.
\newblock \emph{Cognitive Science}, 41:1202--1241.

\bibitem[{Li et~al.(2018)Li, Ji, Du, Li, and Wang}]{li2018textbugger}
Jinfeng Li, Shouling Ji, Tianyu Du, Bo~Li, and Ting Wang. 2018.
\newblock Textbugger: Generating adversarial text against real-world
  applications.
\newblock \emph{arXiv preprint arXiv:1812.05271}.

\bibitem[{Moosavi-Dezfooli et~al.(2016)Moosavi-Dezfooli, Fawzi, and
  Frossard}]{moosavi2016deepfool}
Seyed-Mohsen Moosavi-Dezfooli, Alhussein Fawzi, and Pascal Frossard. 2016.
\newblock Deepfool: a simple and accurate method to fool deep neural networks.
\newblock In \emph{Proceedings of the IEEE CVPR}, pages 2574--2582.

\bibitem[{Mrk{\v{s}}i{\'c} et~al.(2016)Mrk{\v{s}}i{\'c}, S{\'e}aghdha, Thomson,
  Ga{\v{s}}i{\'c}, Rojas-Barahona, Su, Vandyke, Wen, and
  Young}]{mrkvsic2016counter}
Nikola Mrk{\v{s}}i{\'c}, Diarmuid~O S{\'e}aghdha, Blaise Thomson, Milica
  Ga{\v{s}}i{\'c}, Lina Rojas-Barahona, Pei-Hao Su, David Vandyke, Tsung-Hsien
  Wen, and Steve Young. 2016.
\newblock Counter-fitting word vectors to linguistic constraints.
\newblock \emph{arXiv preprint arXiv:1603.00892}.

\bibitem[{Papineni et~al.(2002)Papineni, Roukos, Ward, and
  Zhu}]{Papineni+:2002}
Kishore Papineni, Salim Roukos, Todd Ward, and Wei-Jing Zhu. 2002.
\newblock {BLEU}: a method for automatic evaluation of machine translation.
\newblock In \emph{Proceedings of the 40th ACL}, pages 311--318, Philadelphia,
  Pennsylvania, USA.

\bibitem[{Pennington et~al.(2014)Pennington, Socher, and
  Manning}]{pennington2014glove}
Jeffrey Pennington, Richard Socher, and Christopher Manning. 2014.
\newblock Glove: Global vectors for word representation.
\newblock In \emph{Proceedings of the EMNLP (2014)}, pages 1532--1543.

\bibitem[{Szegedy et~al.(2013)Szegedy, Zaremba, Sutskever, Bruna, Erhan,
  Goodfellow, and Fergus}]{Szegedy+:2013}
Christian Szegedy, Wojciech Zaremba, Ilya Sutskever, Joan Bruna, Dumitru Erhan,
  Ian Goodfellow, and Rob Fergus. 2013.
\newblock Intriguing properties of neural networks.
\newblock \emph{arXiv preprint arXiv:1312.6199}.

\bibitem[{Tsai et~al.(2019)Tsai, Yang, and Chen}]{tsai2019adversarial}
Yi-Ting Tsai, Min-Chu Yang, and Han-Yu Chen. 2019.
\newblock Adversarial attack on sentiment classification.
\newblock In \emph{Proceedings of the 2019 ACL Workshop BlackboxNLP: Analyzing
  and Interpreting Neural Networks for NLP}, pages 233--240.

\bibitem[{Vaswani et~al.(2017)Vaswani, Shazeer, Parmar, Uszkoreit, Jones,
  Gomez, Kaiser, and Polosukhin}]{Vaswani+:2017}
Ashish Vaswani, Noam Shazeer, Niki Parmar, Jakob Uszkoreit, Llion Jones,
  Aidan~N Gomez, Łukasz Kaiser, and Illia Polosukhin. 2017.
\newblock Attention is all you need.
\newblock In \emph{Advances in Neural Information Processing Systems 30}, pages
  5998--6008.

\bibitem[{Wang et~al.(2019)Wang, Wang, Wang, Wang, and Ye}]{wang2019robust}
Wenqi Wang, Lina Wang, Run Wang, Zhibo Wang, and Aoshuang Ye. 2019.
\newblock \href {http://arxiv.org/abs/1902.07285} {Towards a robust deep neural
  network in texts: A survey}.

\bibitem[{Wang et~al.(2016)Wang, Huang, Zhao et~al.}]{wang2016attention}
Yequan Wang, Minlie Huang, Li~Zhao, et~al. 2016.
\newblock Attention-based lstm for aspect-level sentiment classification.
\newblock In \emph{Proceedings of the 2016 conference on empirical methods in
  natural language processing}, pages 606--615.

\bibitem[{Xiao et~al.(2018)Xiao, Li, Zhu, He, Liu, and
  Song}]{xiao2018generating}
Chaowei Xiao, Bo~Li, Jun-Yan Zhu, Warren He, Mingyan Liu, and Dawn Song. 2018.
\newblock Generating adversarial examples with adversarial networks.
\newblock \emph{arXiv preprint arXiv:1801.02610}.

\bibitem[{Yang et~al.(2019)Yang, Dai, Yang, Carbonell, Salakhutdinov, and
  Le}]{Yang+:2019}
Zhilin Yang, Zihang Dai, Yiming Yang, Jaime~G. Carbonell, Ruslan Salakhutdinov,
  and Quoc~V. Le. 2019.
\newblock {XLNet}: Generalized autoregressive pretraining for language
  understanding.
\newblock \emph{CoRR}, abs/1906.08237.

\end{thebibliography}

\end{document}